\documentclass[letterpaper, 10 pt, conference]{ieeeconf}  % Comment this line out if you need a4paper

\IEEEoverridecommandlockouts                              % This command is only needed if 
\usepackage{graphics} % for pdf, bitmapped graphics files
\usepackage{epsfig} % for postscript graphics files
\usepackage{mathptmx} % assumes new font selection scheme installed
\usepackage{times} % assumes new font selection scheme installed
\usepackage{amsmath} % assumes amsmath package installed
\usepackage{amssymb}  % assumes amsmath package installed
\usepackage{multirow}
\usepackage{tabularx}
\usepackage{threeparttable}
\usepackage[final]{changes}

\usepackage[symbol]{footmisc}
\renewcommand*{\thefootnote}{\arabic{footnote}}
\usepackage{scrextend}
\deffootnote[0.8em]{0em}{10em}{\textsuperscript{\thefootnotemark}\,\enskip}

\title{\LARGE \bf
A reinforcement learning path planning approach for range-only underwater target localization with autonomous vehicles
}

\author{Ivan Masmitja$^{1}$, Mario Martin$^{2}$, Kakani Katija$^{3}$, Spartacus Gomariz$^{4}$ and Joan Navarro$^{5}$% <-this % stops a space
\thanks{$^{1}$Ivan Masmitja is with the Bioinspiration Lab, MBARI, Moss Landing CA 95062 USA, and the Institut de Ciències del Mar, CSIC, 08003 Barcelona, Spain {\tt\small masmitja@icm.csic.es}}%
\thanks{$^{2}$Mario Martin is with the Knowledge Engineering and Machine Learning Group, Universitat Politècnica de Catalunya, Barcelona Tech., 08034 Barcelona, Spain {\tt\small mmartin@cs.upc.edu}}%
\thanks{$^{3}$Kakani Katija is with the Bioinspiration Lab, MBARI, Moss Landing, CA 95062, USA {\tt\small kakani@mbari.org}}%
\thanks{$^{4}$Spartacus Gomariz is with the SARTI Research Group, Electronics Department Universitat Politècnica de Catalunya, Barcelona Tech., 080934 Barcelona, Spain. {\tt\small spartacus.gomariz@upc.edu}}%
\thanks{$^{5}$Joan Navarro is with the Institut de Ciències del Mar, CSIC,  08003 Barcelona, Spain {\tt\small joan@icm.csic.es}}%
}

\begin{document}

\maketitle
\thispagestyle{empty}
\pagestyle{empty}

\begin{abstract}
Underwater target localization using range-only and single-beacon (ROSB) techniques with autonomous vehicles has been used recently to improve the limitations of more complex methods, such as long baseline and ultra-short baseline systems. Nonetheless, in ROSB target localization methods, the trajectory of the tracking vehicle near the localized target plays an important role in obtaining the best accuracy of the predicted target position. Here, we investigate a Reinforcement Learning (RL) approach to find the optimal path that an autonomous vehicle should follow in order to increase and optimize the overall accuracy of the predicted target localization, while reducing time and power consumption. To accomplish this objective, different experimental tests have been designed using state-of-the-art deep RL algorithms. Our study also compares the results obtained with the analytical Fisher information matrix approach used in previous studies. The results revealed that the policy learned by the RL agent outperforms trajectories based on these analytical solutions, e.g. the median predicted error at the beginning of the target's localisation is 17$\%$ less. These findings suggest that using deep RL for localizing acoustic targets \replaced{could}{can} be successfully applied to in-water applications that include tracking of acoustically tagged marine animals by autonomous underwater vehicles. This is envisioned as a first necessary step to validate the use of RL to tackle such problems, which could be used later on in a more complex scenarios.
\end{abstract}

%%%%%%%%%%%%%%%%%%%%%%%%%%%%%%%%%%%%%%%%%%%%%%%%%%%%%%%%%%%%%%%%%%%%%%%%%%%%%%%%
\section{Introduction}

One of the main challenges in marine research lies in underwater positioning of underwater features or assets (e.g., marine species \cite{Danovaro2020} or underwater vehicles \cite{Gonzalez2020}). Due to the large attenuation of radio waves in water \cite{Burdic1984}, Global Positioning System (GPS) signals are not suitable for positioning underwater targets. Nonetheless, acoustic signals can fill the underwater communications gap left by radio waves. Acoustic signals have much greater underwater propagation capabilities \cite{Stojanovic2009}, and therefore, a network of nodes or beacons can be deployed and used to localize underwater targets, which may include Autonomous Underwater Vehicles (AUV), benthic rovers, or acoustically tagged organisms.

Unfortunately, underwater acoustic deployments are often complex and highly economically and logistically expensive \cite{Witze2013}. To avoid these inherent issues, different strategies have been developed for tracking underwater targets, moving from the traditional, moored Long Baseline (LBL) systems \cite{Vickery1998}, to GPS Intelligent Buoy (GIB) systems \cite{Alcocer2006}, or most recently range-only and single-beacon (ROSB) methods \cite{Masmitja2019} where a single AUV surveys a marine area to estimate the position of an acoustically tagged target. Range-only methods have different advantages over angle-related localization methods (e.g., Ultra-Short Baseline (USBL) systems \cite{Reis2016}) because it (i) reduces the power consumption and the number of required devices (e.g. an inertial measurement unit), and subsequently the cost and size of the overall system, and (ii) angle measurements are less robust in rough sea conditions compared to range measurements \cite{Ullah2019}, especially if they are used in small platforms \cite{Costanzi2017} such as an Autonomous Surface Vehicle (ASV) Wave Glider (Liquid Robotics, USA).

\begin{figure}[!tb]
\setlength{\fboxsep}{0pt}%
\setlength{\fboxrule}{0pt}%
\centering
\includegraphics[width=0.99\linewidth]{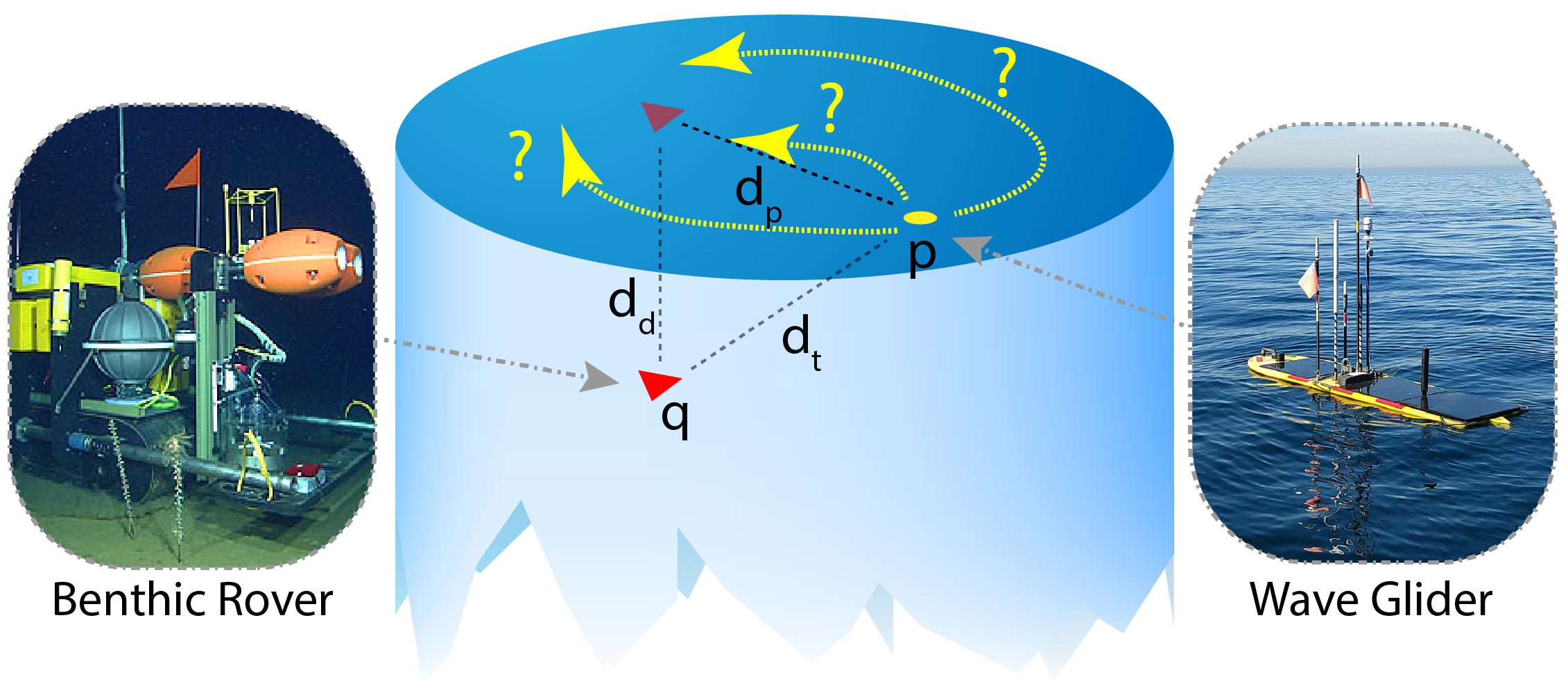}
\caption{What is the optimal path \replaced{that}{than} an agent (yellow dot) should follow in order to track accurately underwater targets (red triangle)? where $q$ is the target's position, $p$ is the agent's position, $d_t$ is the slant range measured, $d_d$ is the target depth, and $d_p$ is distance between the agent and the projected target position into the 2D plane of the agent.}
\label{fig:representation}
\end{figure}

The main drawback in ROSB localization techniques is related to path optimization (i.e. what trajectory should follow the ASV to increase the accuracy of the predicted target position). The ultimate goal is to compute the optimal ASV trajectory that will yield the best possible accuracy of the estimated target positions, which will depend significantly on the trajectories imparted with the ASV. While for static targets the optimization solution is relatively straightforward, in a dynamic environment with a mobile target, an analytical solution is not trivial. In the present work, a deep Reinforcement Learning (RL) approach has been used to find the optimal policy that an agent (e.g., ASV) should follow in order to accurately localize underwater targets (Fig.  \ref{fig:representation}). This is envisioned as a first necessary step to validate the use of RL to tackle such problems, which could be used later on in a more complex scenarios.  
\renewcommand*{\thefootnote}{\fnsymbol{footnote}}
Here, we show that the RL agent\footnote[2]{\textbf{Data and materials availability:} The range-only target localization algorithms with deep RL are available on GitHub: github.com/imasmitja/RLforUTracking} can learn an optimal policy, with a performance comparable to the analytically derived \deleted{optimal }trajectory during the steady state. In addition, our method outperforms the classical strategy of going \replaced{directly}{direct} to the last estimated target position and start to conduct loops around it, with a reduction in predicted error by 17\% during the transient state.
\renewcommand*{\thefootnote}{\arabic{footnote}}

%\textbf{Data and materials availability}: The range-only target localization algorithms with deep RL are available on GitHub: github.com/imasmitja/RLforUTracking.

\section{Related Work}
The relationship between the acoustic sensor location and the accuracy that can be achieved in parameter estimation under different measurement typologies has been widely studied \cite{Ucinski2004}. In general, the computation of the optimal sensor configuration can be carried out analytically by examining the corresponding Cramer-Rao Bound (CRB) or its Fisher Information Matrix (FIM) \cite{Trees2004}. If a set of noisy observations are used to estimate a certain parameter of interest, the CRB sets the lowest bound on the covariance matrix that can asymptotically be achievable using any unbiased estimation algorithm. For example, the CRB method was used to find the optimal sensors’ locations of an underwater sensor network to find a target using their ranges \cite{Moreno-Salinas2016}, and to study the localization accuracy of a target using Time Difference Of Arrivals (TDOA) measurements on different sensor geometry scenarios \cite{Kaune2011}. This approach was adapted to derive the optimal path shape that an ASV should take in order to compute the position of an underwater target using range-only and single-beacon techniques \cite{Masmitja2018}. 

Nevertheless, such analytical studies can be computationally intractable, and especially for multi-target, multi-tracker missions, where a group of autonomous vehicles (trackers) try to localize a set of acoustic-tagged targets. Approaches to address these challenges include using a set of assumptions (e.g., knowing the target maneuvers and knowing one of the tracker's trajectories), or using numerical optimization methods given the complexity of the problem \cite{Crasta2018}. A set of Monte Carlo simulations has also been used to study different triangulation algorithms and derived the optimal path and practices to track underwater moving targets \cite{Masmitja2019}. However, there are scenarios where the information needed to estimate the target position is scarce. For example, in the area-only method \cite{Masmitja2020}, the information used to infer the target position is the area bounded by the maximum range that the pings generated by an acoustic tag can be detected. In such case, the FIM analysis cannot be used to find the optimal sensor placement, and therefore, is even harder to find the path that a tracker should conduct.

Within this framework, we propose the use of deep Reinforcement Learning (RL) techniques to find the optimal trajectory for an Autonomous Surface Vehicle (ASV) to track an underwater target. Deep RL uses the formal framework of Markov Decision Process (MDP) to define the interaction between a learning agent and its environment in terms of states, actions, and rewards \cite{Sutton1998}.

Whereas most of the attention in deep RL has focused on game theory (e.g., to solve Atari games \cite{mnih2013} or to mastering the game of Go \cite{Silver2016}), the same principles can be used to solve path planning and trajectory optimization. Previous studies have shown that gliders can navigate atmospheric thermals autonomously using RL to provide an appropriate framework that identifies an effective navigational strategy as a sequence of decisions made in response to environmental cues \cite{Reddy2018}. \replaced{RL can also be used}{Or} to station a stratospheric Loon superpressure balloon at multiple locations using a RL controller \cite{Bellemare2020}. In addition, a RL algorithm has been trained to efficiently navigate in vortical flow fields \cite{Gunnarson2021}. Finally, an actor-critic architecture has also been used to track a ground target by an unmanned autonomous vehicle \cite{Li2020}, where the RL network is able to control an agent to avoid collisions and reach the target using range and angle information by using Recurrent Neural Networks (RNN). 

In addition, autonomous navigation systems are typically divided into three main layers, which are known as Guidance, Navigation, and Control systems (GNC) \cite{Fossen2002,Masmitja2018b}. The Navigation and Control system strongly depends on platform's configuration and the instruments/sensors used. Here we propose the development of a RL approach as a path planning system (which establishes the points to cross to accomplish the goal of the mission) for an adaptive ASV which will be able to explore the area and locate the detected targets. The algorithm will be designed detached to the lower Control and Navigation layers in order to make it platform-free and easily deployable in real environments.    

\section{Problem Formalization and Notation}
In this paper, we consider the case of a single tracker (an ASV) and a single target (a benthic \added{-seabed-} \replaced{rover}{instrument platform}), hereinafter the agent and the target, respectively. The final goal of the agent is to localize and track the target. Two key algorithms run simultaneously to achieve this goal: (i) agent path planning, which is based on the policy learned using the RL; and (ii) the target position estimation based on range data acquired on-line, where we used a Least Square (LS) approach for its simplicity and low \added{computational time at} runtime \deleted{consumption }\cite{Masmitja2019}. Here, we focused on solving the common scenario where the agent moves in a 2D environment (e.g. an ASV) and the target’s depth is known by the agent. Used for example in \cite{Masmitja2018} and \cite{Smith2021}. Both the agent and the target have an acoustic modem, which can be used to measure the distance between them. Finally, we also assume that the agent knows its position by using their own navigation methods (e.g., GPS or dead reckoning).

\subsection{Environment}
%The target tracking environment is illustrated in Figure \ref{fig:env}. 
The environment is based on OpenAI Particle \cite{Ryan2017, Terry2020}, which is a multi-agent particle world with a continuous observation and action space. This environment has been modified to incorporate the target estimation algorithm (which is based on a LS range-only triangulation technique) and its visualization. The OpenAI Particle action space has been modified to fit the constraints of our scenario that is explained in the following subsections.

%\begin{figure}[!tb]
%\setlength{\fboxsep}{0pt}%
%\setlength{\fboxrule}{0pt}%
%\centering
%\includegraphics[width=0.6\linewidth]{images/environment.jpg}
%\caption{Simulation environment to train the agent, where the agent (blue dot), the target (black dot), and the estimated target position (red dot) is represented. The light blue dots represent the previous agent’s position.}
%\label{fig:env}
%\end{figure}

\subsection{Agent Model}
In the absence of ocean currents the kinematics model of an autonomous vehicle is given by
\begin{equation}
\begin{cases}
&\dot{\textbf{p}}(t) = \textbf{v}(t) \\
&\dot{\textbf{v}}(t) = \textbf{F}/m
\end{cases},
\end{equation} 
where $t \in [0, t_f]; t_f>0$, $\textbf{p} \in \mathbb{R}^2$ is the position vector of the agent in a 2D plane, $\textbf{v} \in \mathbb{R}^2$ is the velocity vector, $\textbf{F} \in \mathbb{R}^2$ is a force vector, and $m$ is the mass of the agent. In this experiment, we have considered an agent with a constant velocity $v$, and a single action space referred to the yaw angle $\psi$. This is a common operational mode when it is applied to torpedo-shape AUVs (e.g., the Tethys LRAUV (MBARI, USA)), or vehicles that does not use thrusters (e.g., the Wave Glider (Liquid Robotics, USA)). Consequently, using a state space formulation, and defining the input action vector $\Delta \psi = u+w \in \mathbb{R}^1$ related to the increment of the agent's angle $\psi$, with zero-mean additive Gaussian noise $w \sim \mathcal{N}(0,\sigma^2)$, the simplified dynamic discrete model at time-step $t$ can be defined by
\begin{equation}
\begin{bmatrix}
\textbf{p}_{t+1} \\ 
\textbf{v}_{t+1} \\
\psi_{t+1} 
\end{bmatrix} = 
\begin{bmatrix}
\textbf{p}_{t} \\ 
0 \\
\psi_{t} 
\end{bmatrix} + 
\begin{bmatrix}
v\textbf{g}(\psi_{t+1})\Delta t \\
v\textbf{g}(\psi_{t+1}) \\
\Delta \psi_t 
\end{bmatrix},
\end{equation} 
where $\textbf{g}(\cdot) \triangleq  [\text{cos}(\cdot),\text{sin}(\cdot)]$, and $\Delta t$ is the sampling time-interval. This equation will set the following way-point to be reached by the agent given a defined time step and agent's velocity. As we stated before, we have designed a path planing algorithm which is detached to the lower Control and Navigation system layers \cite{Fossen2002}, in order to make this method platform free and useful for different autonomous vehicles. Consequently, the action control provided by the RL algorithm is the increment of the yaw angle which the ASVs lower control system should follow using their own internal closed-loop method.

\subsection{Target Model}
In this study, we assumed a static target scenario, and therefore, after its initialization the target position is not changed during all the episode. Thus, the target position vector is defined as $\textbf{q} \in \mathbb{R}^2$.

%Let $\textbf{q} \in \mathbb{R}^2$ be the target position vector. In this study, we have assumed two scenarios: one with a fixed target position, and one with a random walk with Lévy flight distribution \cite{Yang2014}. The simplified dynamic discrete model at time-step $t$ can be obtained as follows
%\begin{equation}
%\begin{bmatrix}
%\textbf{q}_{t+1} \\ 
%\textbf{v}_{t+1}
%\end{bmatrix} = 
%\begin{bmatrix}
%\textbf{q}_{t} \\ 
%\textbf{v}_{t}
%\end{bmatrix} + 
%\begin{bmatrix}
%\textbf{v}_{t+1} \\
%\textbf{u}_{t}/m
%\end{bmatrix} \Delta t,
%\end{equation} 
%where the input vector $\textbf{u}_t$ is equal to 
%\begin{equation}
%\textbf{u}_t = f_t \textbf{g}(\chi_t),
%\end{equation} 
%where $f_t$ is the force applied to the target at time step t, generated following the Lévy distribution, using the so-called Mantegna algorithm for a symmetric Lévy stable distribution, and  $\chi_t \in [0,2 \pi)$ is the angle of the applied force, which is sampled using a random uniform distribution.

\subsection{Measurement Model}
The agent is equipped with a sensor that measures distances to the targets at specified discrete intervals of time. Therefore, the range measurement is naturally modeled in a discrete-time setting as
\begin{equation}
\bar{d}_t = ||\textbf{d}_t||+w_t, \;\;\; t \in \{1,2, \ldots , m\},
\end{equation} 
Where $\textbf{d}_t = \textbf{p}_t - \textbf{q}_t$ is the relative position vector of the target with respect to the agent,  $m$ indicates the number of measurements carried out, and $w_t \sim \mathcal{N}(\epsilon,\sigma^2)$  is a non-zero mean Gaussian measurement error where $\sigma^2$ is the variance and $\epsilon$ is the systematic error, mostly due to the sound speed uncertainty under water \cite{Stojanovic2009,Masmitja2016}. Finally, the projected planar range measurement $d_p$ can be derived knowing the target depth $d_d$ as $\bar{d}_{pt}= \sqrt ({\bar{d}_t}^2 - {d_d}^2)$.

\subsection{Target Position Prediction Model}
Different methods can be used to obtain an estimation of the target's position $\hat{\textbf{q}}$ using range-only and single-beacon techniques \cite{Masmitja2019}. Here, a simple unconstrained LS algorithm is used. The main idea on LS algorithms lies in a linearisation of the system by using the squared range measurements to obtain a linear equation as a function of the unknown target’s position.
While this technique is suitable for static target localization, its capability to track a moving target can be compromised, falling compared to other algorithms (e.g., Particle Filter (PF)). However, the run-time performance of the LS is orders of magnitudes below its competitors, which is key in RL techniques to accelerate the training phase.

\subsection{Observation and Action Space}
The observations at each time-step $t$ that we can get from the environment include the position $\textbf{p}$ and velocity $\textbf{v}$ vectors of the agent, the relative position vector of the estimated target position ($\hat{\textbf{d}}_t = \textbf{p}_t - \hat{\textbf{q}}_t$), and the projected distance measured by the sensor $\bar{d}_{pt}$:
\begin{equation}
\textbf{o}_t = [\textbf{p}_t, \textbf{v}_t, \hat{\textbf{d}}_t,\bar{d}_{pt}].
\end{equation} 
On the other hand, the action space is determined by the force applied to the yaw ($\psi$) angle of the agent, as $a_t \triangleq u_{\psi}$.

\subsection{Reward Function}
In RL, the agent obtains rewards as a function of the state and agent’s action. The agent aims to maximize the total expected return 
$R = \sum_{t=0}^{T} \gamma^t r^t$, where $\gamma$ is a discount factor and $T$ is the time horizon.

The design of a good reward function is a key aspect in RL. In dense reward settings, the agent receives diverse rewards in most states (e.g., a reward proportional to distance to the goal), which allow the agent to quickly differentiate good states from bad ones. However, such \added{an} approach can easily exploit badly designed rewards, and get stuck in local optima and induce behavior that the designer did not intend. In contrast, goal-based sparse rewards are appealing since they do not suffer from the reward exploration problem \cite{Memarian2021}. In addition, this simple small set of rules have its similarities with biological behaviours, and therefore, applicable to animals with very limited level of information processing \cite{Nuzhin2021}.   
 
Here, we propose a combination of both reward methods: (i) a non-sparse reward to guide the agent towards the goal when its performance is poor, and (ii) a sparse reward when the performance of the agent reaches a predefined threshold. In addition, we have defined two different goals to optimize the agent’s trajectory, which influence the reward obtained by the agent: (i) a reward function based on the distance between the agent and the target, and (ii) a reward function based on the estimated target position error.

The reward as a function of the distance between the agent and the target is designed as
\begin{equation} \label{eq:rd}
r_d = \left\{\begin{array}{cll}
\lambda(0.5-\hat{d}) & if &\hat{d}>d_{th}\\
1 & \mbox{else}&\end{array}\right.,
\end{equation} 
where $\lambda$ is a positive constant, $\hat{d}$ is the distance between the agent and the estimated target position, and $d_{th}$  is the predefined distance threshold to be reached by the agent. The smaller the distance $\hat{d}$ is, the closer the agent is to the estimated target, and therefore, this reward is the most important reward to guide the agent navigate to the target.

The reward as a function of the predicted target error is designed as
\begin{equation} \label{eq:re}
r_e = \left\{\begin{array}{cll}
\lambda(0.5-e_q) & if &e_q>e_{th}\\
1 & \mbox{else}&\end{array}\right.,
\end{equation} 
where $e_q= || \hat{\textbf{q}}_t - \textbf{q}_t ||$ is the error between the predicted target position and the real target position at time-step $t$, and $e_{th}$ is the predefined error threshold to be reached by the agent. This reward is the most important to optimize the agent’s trajectory toward the goal of finding the optimal path which leads to the greatest accuracy of the estimated target position.

Finally, a terminal reward related to the success of the mission, has been designed as
\begin{equation}
r_{terminal} = \left\{\begin{array}{cll}
-100 & if & \hat{d} > \hat{d}_{max}\\
-1 & if & \hat{d} < \hat{d}_{min}\\
\phantom{-} 0 & \mbox{else}&\end{array}\right.,
\end{equation} 
where $\hat{d}_{max}$ is the maximum distance where the agent can go related to the target, and $\hat{d}_{min}$ is a threshold set to avoid collisions between the target and the agent. Consequently, this sparse reward gives a higher penalty if the distance between the target and the agent is bigger than a maximum or less than a minimum threshold.

Then, the final reward is given by $r=  r_d+r_e+r_{terminal}$.

\section{Algorithm}

Three different actor-critic algorithms have been implemented and tested to compare their performance:
\begin{itemize} 
\item \textit{Deep Deterministic Policy Gradient (DDPG)}: This deep Q-learning algorithm is an actor-critic, model-free algorithm based on the deterministic policy gradient that can operate over continuous action spaces \cite{Lillicrap2019}.  
\item \textit{Twin Delayed Deep Deterministic Policy Gradient (TD3)}: The TD3 \cite{Fujimoto2018} is a variant of the DDPG, which address the overestimation problem in Actor-Critic methods. Specifically, TD3 employs two critics $Q_1$  and $Q_2$ (with a policy update delay equal to 2), and uses the minimum of the predicted optimal future return in observation $\textbf{o}_{t+1}$ to bootstrap the Q-value of the current observation $\textbf{o}_t$ and action $a_t$. 
\item \textit{Soft Actor-Critic (SAC)}: Model-free deep reinforcement learning algorithms typically suffer from two major challenges, very high sample complexity and brittle convergence properties, which necessitate meticulous hyperparameter tuning. However, in this off-policy actor-critic deep RL algorithm, the actor aims to maximize expected reward while also maximizing entropy \cite{Haarnoja2018}. That is, to succeed at the task while acting as randomly as possible.
\end{itemize}

%A top level representation of the structure of these recurrent actor-critic framework is illustrated in %Figure \ref{fig:structure}.
%\begin{figure}[!tb]
%\setlength{\fboxsep}{0pt}%
%\setlength{\fboxrule}{0pt}%
%\centering
%\includegraphics[width=0.8\linewidth]{images/structure.jpg}
%\caption{Overview of the actor-critic framework.}
%\label{fig:structure}
%\end{figure}

\section{Results}
A set of trials has been conducted to evaluate deep RL algorithms as a guidance system for an ASV. The results showed the performance obtained with the learned policy to localise a static target using ROSB triangulation techniques. In addition, it has been compared against the optimal trajectory derived analytically \cite{Masmitja2018}, which is a set of measurements equally distributed on a circumference centred on top of the target (hereinafter referred to as \textit{predefined path}). With a circumference's radius at least equal to $\sqrt{2}$ multiplied by the depth of the target, or greater.

\subsection{Experiment Settings}
The following hyperparameters and environment settings have been used during the training (Table \ref{table:hyper}). The agent's constant velocity was set to $v=1$ m/s and the sampling time interval to $\Delta t = 30$ s. In addition, all the distances were scaled to 1, which represented an horizon of 1 km. The reward was initialized with a $\lambda = 0.01$, which was empirically found as \replaced{a good}{an optimal} value, and the $d_{th}$ in (\ref{eq:rd}) to 300 m. The measurement noise $w_t$ was set with a $\sigma$ of 1 m and $\epsilon$ of $1\%$ of the distance\deleted{, which is a value close to real conditions}. Finally the number of steps per episode was set to 200.

\begin{table}[ht!]
\begin{threeparttable}[b]
\caption{Hyperparameters for algorithms}
\label{table:hyper}
\centering
\begin{tabularx}{0.99\linewidth} { 
   >{\raggedright\arraybackslash}X 
   >{\centering\arraybackslash}c 
   >{\centering\arraybackslash}c 
   >{\centering\arraybackslash}c }
%\begin{tabularx}{0.95\linetwidth}{bss}
\hline
\multirow{2}{4em}{\textbf{Hyperparameter}} & \multicolumn{3}{c}{\textbf{Algorithms}} \\
\cline{2-4}
& \textbf{DDPG} & \textbf{TD3} & \textbf{SAC} \\
\hline
Replay buffer size ($D$) & \multicolumn{3}{c}{500000}\\
\hline
Batch size ($N$) & \multicolumn{3}{c}{32$^*$}\\
\hline
Discount factor ($\gamma$) & \multicolumn{3}{c}{0.99}\\
\hline
Target NN update rate ($\tau$) & \multicolumn{3}{c}{0.01}\\
\hline
Actor learning rate ($l_a$) & \multicolumn{3}{c}{1e-3}\\
\hline
Critic learning rate ($l_c$) & \multicolumn{3}{c}{1e-4}\\
\hline
Optimizer  & \multicolumn{3}{c}{Adam}\\
\hline
Random start episode number  & \multicolumn{3}{c}{10000}\\
\hline
Update every & \multicolumn{3}{c}{30}\\
\hline
Update times & \multicolumn{3}{c}{20}\\
\hline
Parallel envs & \multicolumn{3}{c}{8}\\
\hline
Actor NN structure & \multicolumn{3}{c}{[64,32]}\\
\hline
Critic NN structure & \multicolumn{3}{c}{[64,32]}\\
\hline
Actor exploration noise & \multicolumn{2}{c}{0.5} & -\\
\hline
Noise reduction per episode & \multicolumn{2}{c}{0.9999} & -\\
\hline
Policy update delay & - & 2 & -\\
\hline
Entropy regulation coefficient ($\alpha$)  & - & - & $0.005^{\sharp}$ \\ 
\hline
\end{tabularx}
\begin{tablenotes}
       \item [$*$] increases by 2 every 200000 episodes, up to 2048, as in \cite{Smith2018}.
       \item [$\sharp$] SAC is also configured with an automatic entropy regulation using Adam optimizer.
     \end{tablenotes}
  \end{threeparttable}
\end{table}

\subsection{Experimental Results for Static Targets}
One of the key questions is to see if an agent can find the optimal policy to localise an underwater target using the range-only method. We tested 3 different reward function configurations based on the predicted target error $e_q$ (\ref{eq:re}) :
\begin{itemize}
\item Test 1: $e_{th} = 0$ m (Non-sparse reward)
\item Test 2a: $e_{th} = 1$ m (Non-sparse + Sparse reward)
\item Test 2b: $e_{th} = 0.3$ m (Non-sparse + Sparse reward)
\end{itemize}

\begin{figure*}[tbhp]
\setlength{\fboxsep}{0pt}%
\setlength{\fboxrule}{0pt}%
\centering
\includegraphics[width=0.95\textwidth]{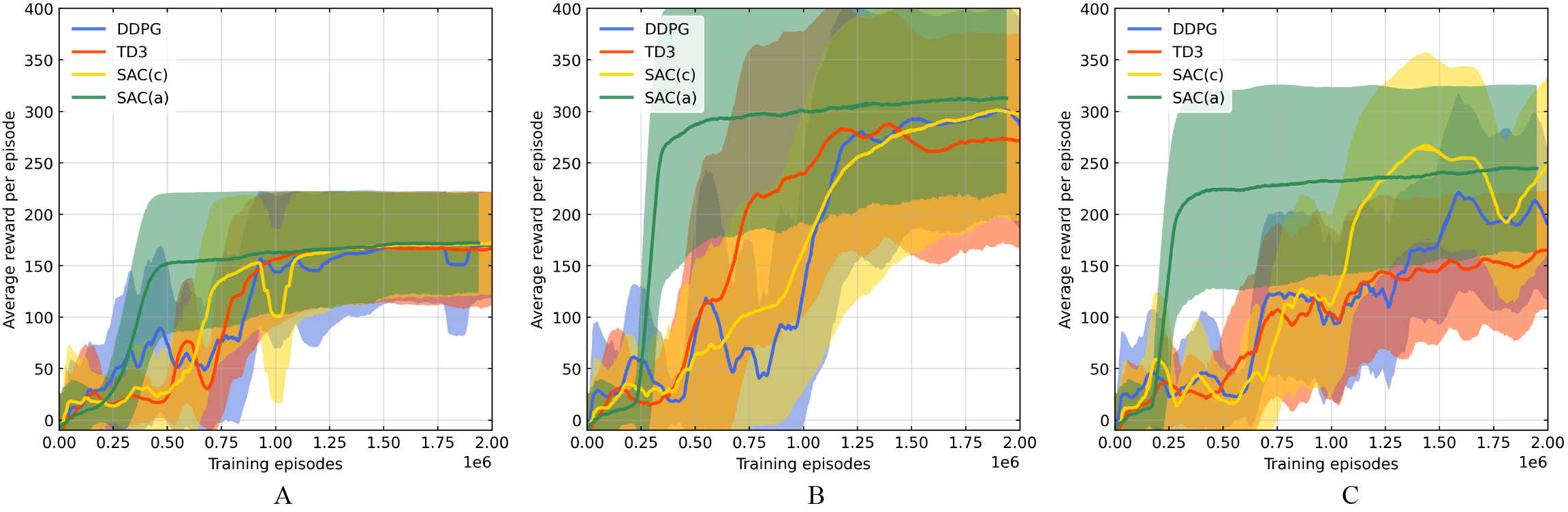}
\caption{Average reward and SD per episode obtained using the DDPG, TD3, and SAC algorithms, where SAC(c) indicates a constant entropy regularization parameter, and SAC(a) indicates an automatic entropy regulation using the Adam optimizer. Average obtained using a rolling window of the latest 100000 episodes. A single agent has been trained using three different reward functions: (A) Test 1, a non-sparse reward; (B) Test 2a, a non-sparse + sparse reward; and (C) Test 2b, a non-sparse + sparse reward with a more constrained error threshold.}
\label{fig:comp_rew}

\bigskip

\includegraphics[width=0.95\textwidth]{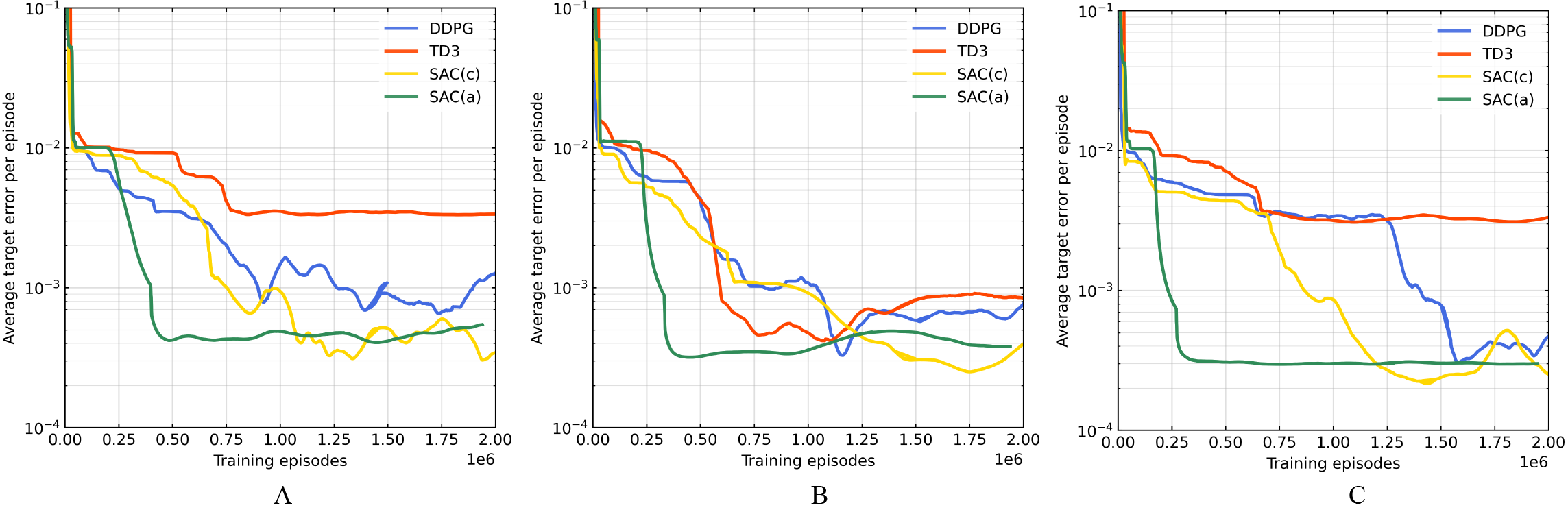}
\caption{Average target error per episode obtained using the DDPG, TD3, and SAC algorithms. Average obtained using a rolling window of the latest 10000 episodes. A single agent has been trained using three different reward functions: (A) Test 1, a non-sparse reward; (B) Test 2a, a non-sparse + sparse reward; and (C) Test 2b, a non-sparse + sparse reward with a more constrained error threshold.}
\label{fig:comp_err}

\bigskip

\includegraphics[width=0.95\textwidth]{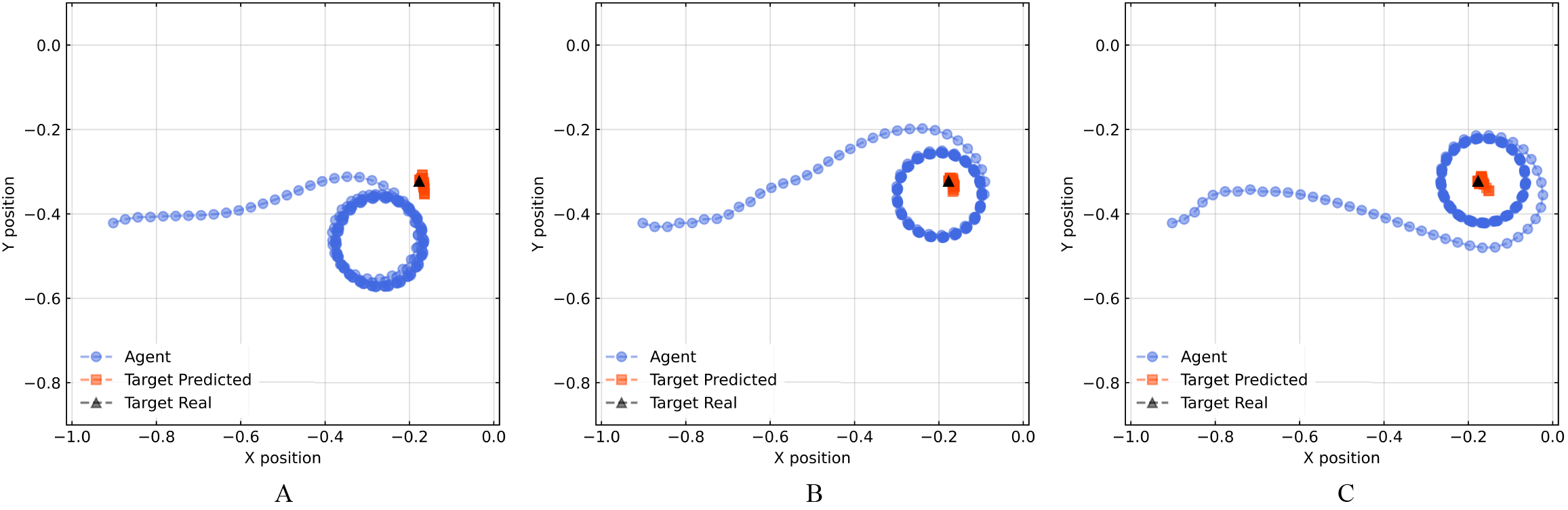}
\caption{A single trajectory of a trained agent \added{(distances in km)}. Blue dots are the trajectory of the agent, where each dot indicates where a new update was conducted (i.e. a new range measured, an update of the estimated target position, and a new action chosen). Red squares are the estimated target position. (A) TD3 with reward equal to Test 1, (B) TD3 with reward equal to Test 2a, and (C) SAC(c) with reward equal to Test 2b.}
\label{fig:comp_trained}
\end{figure*}

We \replaced{measured}{found} the average reward and Standard Deviation (SD) per episode obtained during the training (Fig.  \ref{fig:comp_rew}). The three algorithms implemented (DDPG, TD3, and SAC) have been tested using the different reward function configurations explained above. The average reward and SD per episode has been obtained using a rolling window of the latest 100000 episodes. We can see that the SAC(a) out-performed the rest of the algorithms in all the different reward functions. 

While the average reward has information related to the accuracy of the predicted target position, it is difficult to compare the results and have an idea of which reward function configuration gives the greatest performance. Therefore, we computed the average predicted target error per episode (using a rolling window of the latest 10000 episodes; Fig \ref{fig:comp_err}), which gives us a clear metric for what configuration can achieve the greatest accuracy on estimating the target position. We found a small variation in the average reward per episode obtained between SAC and TD3 algorithms on Test 1 (Fig \ref{fig:comp_err}A), and the highest accuracy was obtained using the SAC(c) under the reward function of Test 2b, whereas the more stable one was the SAC(a).

We can see the trajectories conducted by the agents trained under the three configurations of the reward function in Fig.  \ref{fig:comp_trained}. Here we used the agent trained with the TD3 algorithm in the first two reward functions and the SAC(c) algorithm in the last one. This was done because the TD3 has greater variability among the reward functions designed, and the SAC is the one that presents the greatest performance. In summary, these plots reveal interesting behaviours learned by the agents:
\begin{itemize}
\item TD3 with reward equal to Test 1 (Fig.  \ref{fig:comp_trained}A): The agent learns to go close to the target, but it conducts loops outside the position of the target (i.e. the distance between the center of the loops, conducted by the agent, and the target is greater than the radius of the loops themselves). This type of behaviour is known to perform poorest related to the accuracy of the estimated target positions \cite{Masmitja2019}. Nonetheless, the agent is always inside the 300 m boundary delimited by $d_{th}$, and therefore, the reward $r_d$ obtained is maximized. In addition, because the reward $r_e$ is much less compared to $r_d$, the agent can not learn its exploitation. 
\item TD3 with reward equal to Test 2a (Fig.  \ref{fig:comp_trained}B): In this case, the agent has learned to conduct loops centered on top of the agent but with some offset (i.e. the distance between the center of the loops conducted by the agent and the target is less than the loops radius, but greater than 0). This behaviour increases the accuracy of the estimated target positions. While this behaviour was learned by increasing the reward $r_e$ when the agent reached a certain accuracy threshold $e_{th}$, the agent will not reduce the target localisation error further because the accuracy is below the $e_{th}$. 
\item SAC(c) with reward equal to Test 2b (Fig.  \ref{fig:comp_trained}C): In this more restricted reward configuration (i.e. a lower $e_{th}$), the TD3 cannot exploit correctly the reward function and it reached a sub-optimal policy. Nonetheless, the agent trained with a SAC(c) algorithm, has learned the \textit{predefined path}, which is to conduct loops centered on top of the target with nearly a zero offset.
\end{itemize}

We also see that the agent has learned a \textit{sinusoidal} trajectory when it approaches the target (i.e. transient state). Interestingly, \replaced{collinear}{co-lineal} points have a deficient performance when estimating the position of the target using range-only triangulation techniques \cite{Song1999, Jouffroy2006}. Therefore, this behaviour helps the agent to obtain a greater estimation of the target position at the beginning of the experiment. 

\subsection{Policy learning is dependent on target depth}
\begin{figure}[!tbh]
\setlength{\fboxsep}{0pt}%
\setlength{\fboxrule}{0pt}%
\centering
\includegraphics[width=0.67\linewidth]{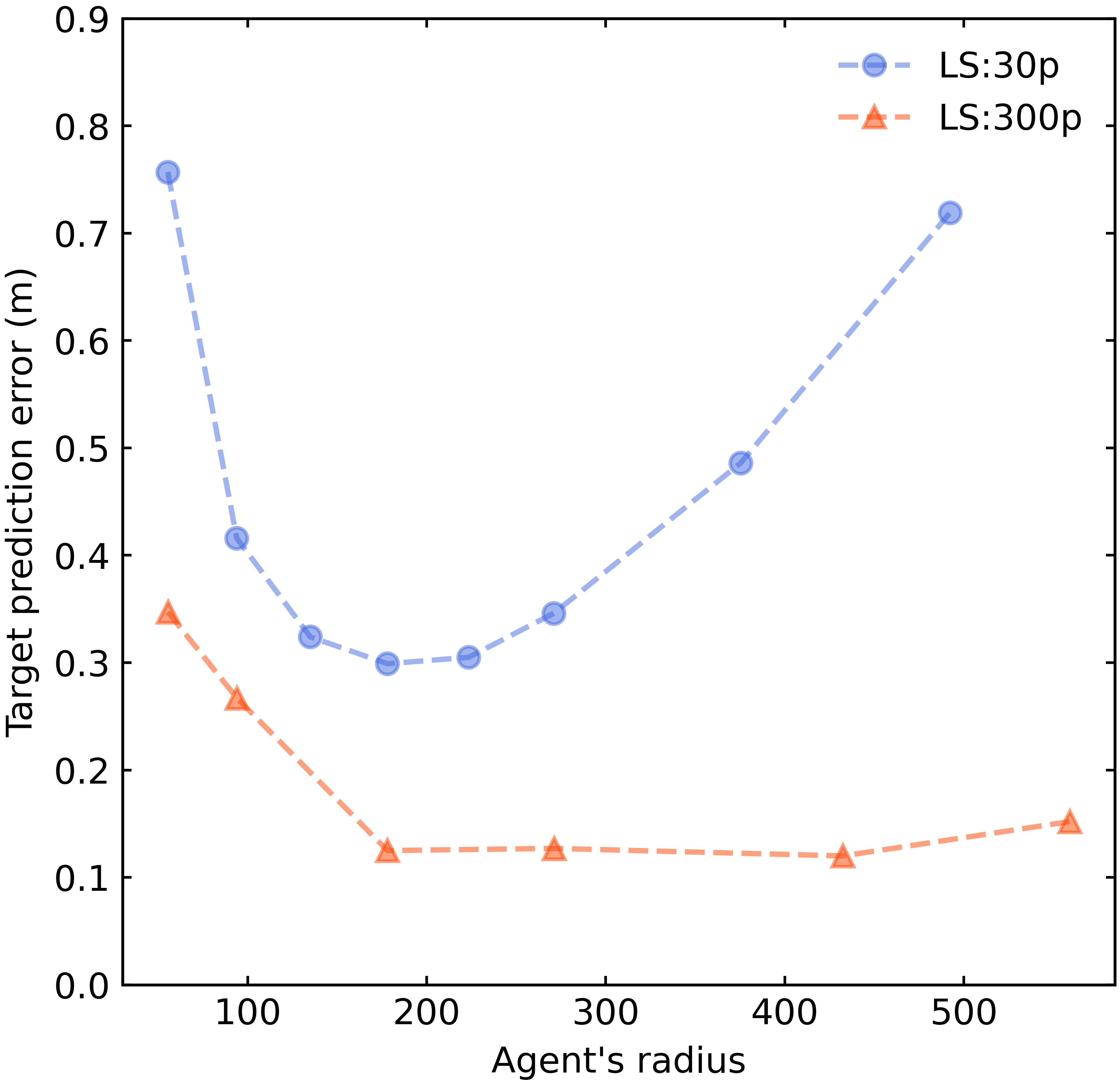}
\caption{Target prediction error as a function of the loop radius conducted by the agent. Trial conducted with a target's depth equal to 200 m. Using both 30 and 300 points to estimate its position by a LS triangulation method.}
\label{fig:depth_ideal}
\end{figure}

The target depth has an influence on sensor placement for range-only target localization in a planar 2D scenario due to the measurement noise $w_t$ and the projection of the slant range into the plane where these measurements where conducted \cite{Moreno2010}. Typically, the location of the measurements have to increase proportional with the target depth in order to maintain or increase the predicted accuracy (i.e. the ideal radius of the circumferential path or loop has to be typically as large as possible). This behaviour can be observed on Fig.  \ref{fig:depth_ideal}, where the target's depth was set to 200 m. Limitations to this method include: (i) the time/power required for the ASV to conduct such large maneuvers, which is even critical for tracking moving targets; or (ii) the number of range measurements required to complete the loop. For example, if the ASV conducts a range measurement every 30 m, and we only use the latest 30 points to estimate the target position using LS. In this case, if the loop's radius is too large, those 30 measurement points will laid only in one side of the circle, which will yield in a bad target prediction (Fig.  \ref{fig:depth_ideal} with an agent's radius $>$ 200 m). 

\begin{figure}[!tbh]
\setlength{\fboxsep}{0pt}%
\setlength{\fboxrule}{0pt}%
\centering
\includegraphics[width=0.69\linewidth]{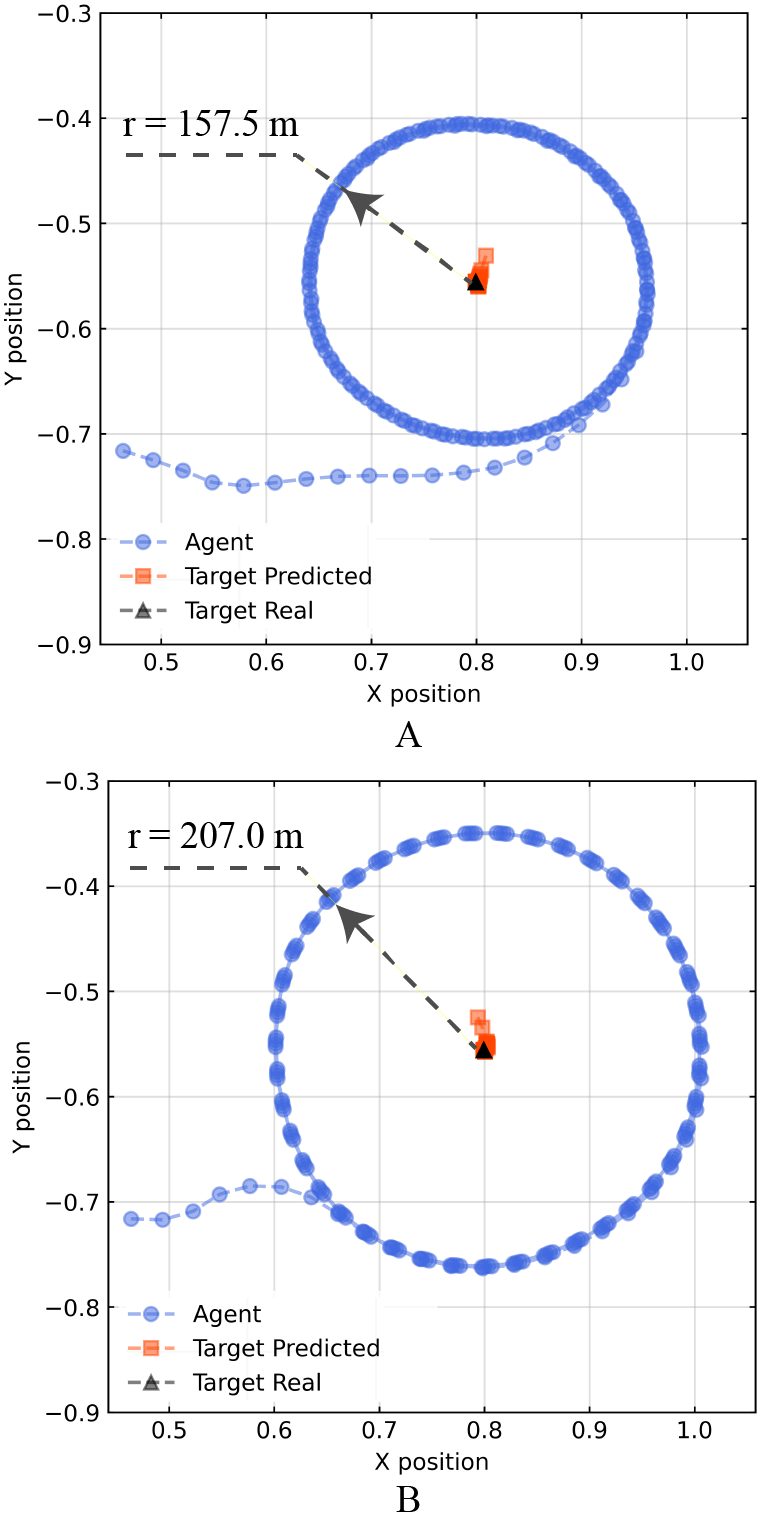}
\caption{Two agents trained at different target depths. The agent has learned to increase the radius of the path as the depth of the target increases from (A) 15 m to (B) 200 m.}
\label{fig:comp_radi}
\end{figure}

We tested our deep RL algorithms under different discrete target depth configurations to determine whether the agents could learn this behaviour and adapt the radius of the loop trajectory with respect to the target's depth. We observed that the SAC(a) agent was able to adapt its trajectory (Fig.  \ref{fig:comp_radi}). With this policy, the error of the predicted target position can be maintained below the 0.16 m threshold.

\subsection{Comparison with a \textit{predefined path}}
\begin{figure}[!bht]
\setlength{\fboxsep}{0pt}%
\setlength{\fboxrule}{0pt}%
\centering
\includegraphics[width=0.7\linewidth]{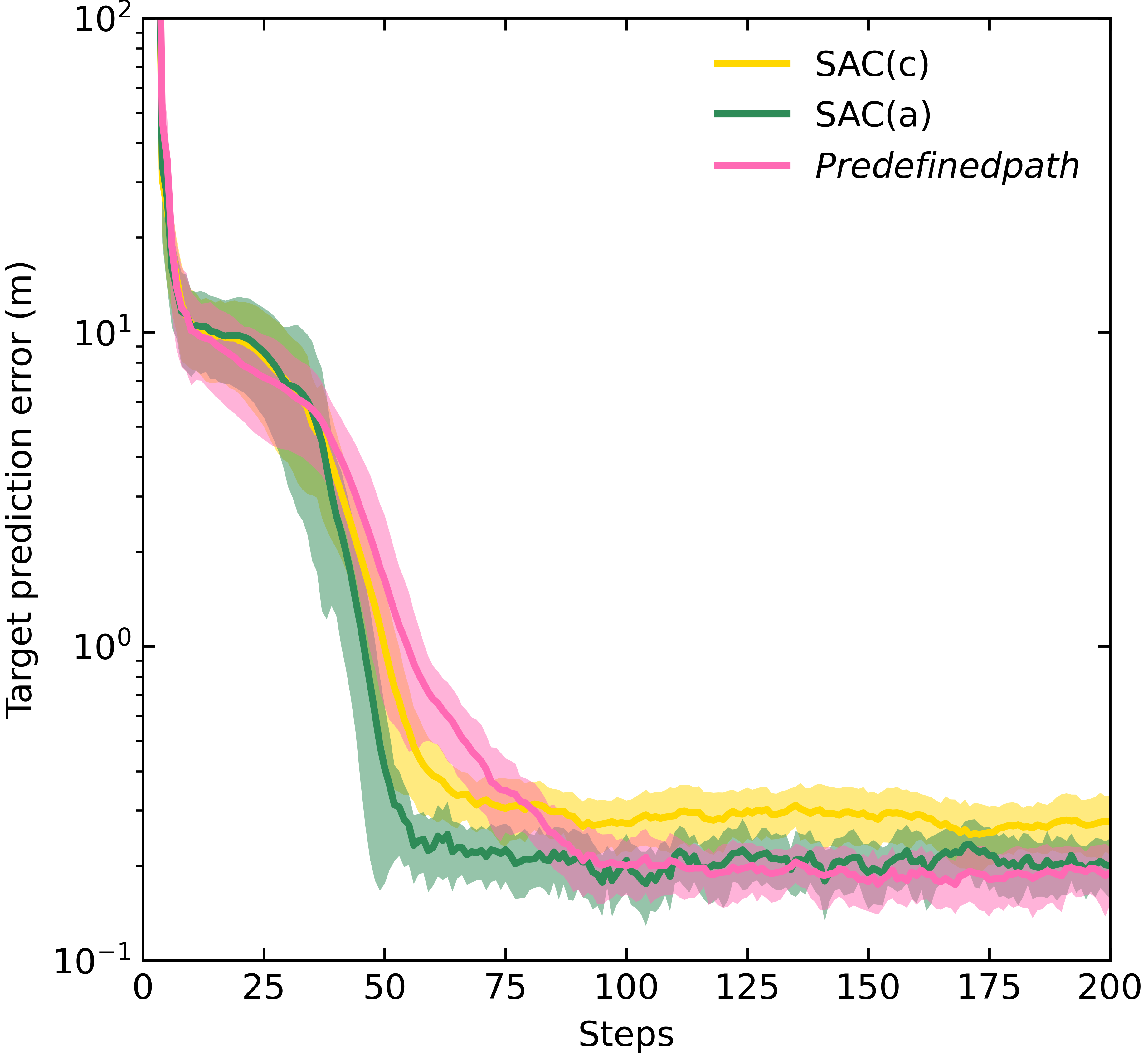}
\caption{Comparison between SAC algorithms and a \textit{predefined path}. Target RMSE prediction error evolution from the first 200 steps and 100 random iterations. Using the Interquartile Mean (IQM) as suggested by \cite{Agarwal2021}. 
%Median and interquartile ranges (IQR) of the RMSE using the first 75 steps (B), and using the last 75 steps (C).
}
\label{fig:comp_results}
\end{figure}

Finally, the performance of the trained agent has been compared with the \textit{predefined path} (aka a loop with a constant radius around the predicted target position) following the reliable evaluation procedures reported in \cite{Agarwal2021}. This trial has been conducted 100 times in the simulation environment for each algorithm: SAC(c), SAC(a), and \textit{predefined path}. Both RL agents has been trained using the reward of Test 2b. The environment used a random seed for each execution, and the range measurement noise $w_t$ was set with a $\sigma$ of 1 m and $\epsilon$ of $1\%$ of the distance as during the agent's training. 
The result shows the evolution of the target estimation over 200 steps, where its Interquartile Mean (IQM) and the SD of the Root Mean Square Error (RMSE) are presented (Fig.  \ref{fig:comp_results}). At the start of tracking (transient state), the SAC algorithm is able to more accurately localise the target. In this case, the average IQM error of SAC(a) at the beginning of the trial is 17$\%$ less than  the \textit{predefined path}, which indicates a probability of improvement equal to 0.61. Finally, at the end of the trial (steady state), SAC(a) has a similar performance to the \textit{predefined path}, which means that the RL agent has learned a policy close to the optimum one derived analytically \cite{Masmitja2018}.

\section{Conclusions}
We demonstrate how deep reinforcement learning can learn optimal trajectories to guide an autonomous vehicle to localize \replaced{a stationary underwater target}{underwater targets}. It is worth noticing that this is envisioned as a first necessary step to validate the use of deep RL to tackle such problems, which could be used later on in more complex scenarios. In the future, the architecture developed here could also be used to train an agent to follow moving underwater assets, and also to train multi-agent and multi-target scenarios, where a group of coordinated agents can navigate to find and track a series of underwater assets at previously unknown positions. This kind of capability opens a new way to deploy adaptive underwater vehicles in a coordinated fashion that are capable of adapting their behaviour to more effectively localize underwater targets.

%\textbf{Do not} include acknowledgements in the initial version of
%the paper submitted for blind review.

% Acknowledgements should only appear in the accepted version.
\section*{Acknowledgements}

%\textbf{Do not} include acknowledgements in the initial version of
%the paper submitted for blind review.

This project has received funding from the European Union’s Horizon 2020 research and innovation programme under the Marie Skłodowska-Curie grant agreement No 893089. This work also received financial support from the Spanish Ministerio de Economía y Competitividad (SASES: RTI2018-095112-B-I00; BITER-ECO: PID2020-114732RB-C31). This work acknowledges the ‘Severo Ochoa Centre of Excellence’ accreditation (CEX2019-000928-S), and from the Generalitat de Catalunya ”Sistemas de Adquisición Remota de datos y Tratamiento de la Información en el Medio Marino (SARTI-MAR)” 2017 SGR 376. We gratefully acknowledge the David and Lucile Packard Foundation. 

%12
\addtolength{\textheight}{-1cm}   % This command serves to balance the column lengths
                                  % on the last page of the document manually. It shortens
                                  % the textheight of the last page by a suitable amount.
                                  % This command does not take effect until the next page
                                  % so it should come on the page before the last. Make
                                  % sure that you do not shorten the textheight too much.
                                  
%\section{References}
{\small
\bibliographystyle{IEEEtran}
\bibliography{IEEEabrv,egbib}

% Generated by IEEEtran.bst, version: 1.14 (2015/08/26)
\begin{thebibliography}{10}
\providecommand{\url}[1]{#1}
\csname url@samestyle\endcsname
\providecommand{\newblock}{\relax}
\providecommand{\bibinfo}[2]{#2}
\providecommand{\BIBentrySTDinterwordspacing}{\spaceskip=0pt\relax}
\providecommand{\BIBentryALTinterwordstretchfactor}{4}
\providecommand{\BIBentryALTinterwordspacing}{\spaceskip=\fontdimen2\font plus
\BIBentryALTinterwordstretchfactor\fontdimen3\font minus
  \fontdimen4\font\relax}
\providecommand{\BIBforeignlanguage}[2]{{%
\expandafter\ifx\csname l@#1\endcsname\relax
\typeout{** WARNING: IEEEtran.bst: No hyphenation pattern has been}%
\typeout{** loaded for the language `#1'. Using the pattern for}%
\typeout{** the default language instead.}%
\else
\language=\csname l@#1\endcsname
\fi
#2}}
\providecommand{\BIBdecl}{\relax}
\BIBdecl

\bibitem{Danovaro2020}
R.~Danovaro, E.~Fanelli, J.~Aguzzi, D.~Billett, L.~Carugati, C.~Corinaldesi,
  A.~Dell’Anno, K.~Gjerde, A.~J. Jamieson, S.~Kark, C.~McClain, L.~Levin,
  N.~Levin, E.~Ramirez-Llodra, H.~Ruhl, C.~R. Smith, P.~V.~R. Snelgrove,
  L.~Thomsen, C.~L. Van~Dover, and M.~Yasuhara, ``Ecological variables for
  developing a global deep-ocean monitoring and conservation strategy,''
  \emph{Nature Ecology and Evolution}, vol.~4, no.~9, pp. 181--192, 2020.

\bibitem{Gonzalez2020}
J.~González-García, A.~Gómez-Espinosa, E.~Cuan-Urquizo, L.~G.
  García-Valdovinos, T.~Salgado-Jiménez, and J.~A.~E. Cabello, ``Autonomous
  underwater vehicles: Localization, navigation, and communication for
  collaborative missions,'' \emph{Applied Sciences}, vol.~10, no.~4, p. 1256,
  2020.

\bibitem{Burdic1984}
W.~S. Burdic and J.~F. Bartram, ``Underwater acoustic system analysis by
  william s. burdic,'' \emph{The Journal of the Acoustical Society of America},
  vol.~76, no.~3, pp. 996--996, 1984.

\bibitem{Stojanovic2009}
M.~Stojanovic and J.~Preisig, ``Underwater acoustic communication channels:
  Propagation models and statistical characterization,'' \emph{IEEE
  Communications Magazine}, vol.~47, no.~1, pp. 84--89, 2009.

\bibitem{Witze2013}
A.~Witze, ``Marine science: Oceanography's billion-dollar baby,''
  \emph{Nature}, vol. 501, no. 7468, pp. 480--482, 2013.

\bibitem{Vickery1998}
K.~Vickery, ``Acoustic positioning systems. a practical overview of current
  systems,'' in \emph{Proceedings of the 1998 Workshop on Autonomous Underwater
  Vehicles (Cat. No.98CH36290)}, 1998, pp. 5--17.

\bibitem{Alcocer2006}
A.~Alcocer, P.~Oliveira, and A.~Pascoal, ``Underwater acoustic positioning
  systems based on buoys with gps,'' in \emph{Proceedings of the Eighth
  European Conference on Underwater Acoustics}, vol.~8, 2006, pp. 1--8.

\bibitem{Masmitja2019}
I.~Masmitja, S.~Gomariz, J.~Del-Rio, B.~Kieft, T.~O’Reilly, P.-J. Bouvet, and
  J.~Aguzzi, ``Range-only single-beacon tracking of underwater targets from an
  autonomous vehicle: From theory to practice,'' \emph{IEEE Access}, vol.~7,
  pp. 86\,946--86\,963, 2019.

\bibitem{Reis2016}
J.~Reis, M.~Morgado, P.~Batista, P.~Oliveira, and C.~Silvestre, ``Design and
  experimental validation of a usbl underwater acoustic positioning system,''
  \emph{Sensors}, vol.~16, no.~9, 2016.

\bibitem{Ullah2019}
I.~Ullah, J.~Chen, X.~Su, C.~Esposito, and C.~Choi, ``Localization and
  detection of targets in underwater wireless sensor using distance and angle
  based algorithms,'' \emph{IEEE Access}, vol.~7, pp. 45\,693--45\,704, 2019.

\bibitem{Costanzi2017}
R.~Costanzi, N.~Monnini, A.~Ridolfi, B.~Allotta, and A.~Caiti, ``On field
  experience on underwater acoustic localization through usbl modems,'' in
  \emph{OCEANS 2017 - Aberdeen}, 2017, pp. 1--5.

\bibitem{Ucinski2004}
D.~Ucinski, \emph{Optimal measurement methods for distributed parameter system
  identification}.\hskip 1em plus 0.5em minus 0.4em\relax CRC press, 2004.

\bibitem{Trees2004}
H.~L. Van~Trees, K.~L. Bell, and Z.~Tian, \emph{Detection, estimation, and
  modulation theory, part I: detection, estimation, and linear modulation
  theory}.\hskip 1em plus 0.5em minus 0.4em\relax John Wiley \& Sons, 2004.

\bibitem{Moreno-Salinas2016}
D.~Moreno-Salinas, A.~Pascoal, and J.~Aranda, ``Optimal sensor placement for
  acoustic underwater target positioning with range-only measurements,''
  \emph{IEEE Journal of Oceanic Engineering}, vol.~41, no.~3, pp. 620--643,
  2016.

\bibitem{Kaune2011}
R.~Kaune, J.~Hörst, and W.~Koch, ``Accuracy analysis for tdoa localization in
  sensor networks,'' in \emph{14th International Conference on Information
  Fusion}, 2011, pp. 1--8.

\bibitem{Masmitja2018}
I.~Masmitja, S.~Gomariz, J.~Del-Rio, B.~Kieft, T.~O’Reilly, P.-J. Bouvet, and
  J.~Aguzzi, ``Optimal path shape for range-only underwater target localization
  using a wave glider,'' \emph{The International Journal of Robotics Research},
  vol.~37, no.~12, pp. 1447--1462, 2018.

\bibitem{Crasta2018}
N.~Crasta, D.~Moreno-Salinas, A.~Pascoal, and J.~Aranda, ``Multiple autonomous
  surface vehicle motion planning for cooperative range-based underwater target
  localization,'' \emph{Annual Reviews in Control}, vol.~46, pp. 326--342,
  2018.

\bibitem{Masmitja2020}
I.~Masmitja, J.~Navarro, S.~Gomariz, J.~Aguzzi, B.~Kieft, T.~O’Reilly,
  K.~Katija, P.~J. Bouvet, C.~Fannjiang, M.~Vigo, P.~Puig, A.~Alcocer,
  G.~Vallicrosa, N.~Palomeras, M.~Carreras, J.~del Rio, and J.~B. Company,
  ``Mobile robotic platforms for the acoustic tracking of deep-sea demersal
  fishery resources,'' \emph{Science Robotics}, vol.~5, no.~48, p. eabc3701,
  2020.

\bibitem{Sutton1998}
R.~S. Sutton and A.~G. Barto, \emph{Reinforcement Learning: An Introduction},
  2nd~ed.\hskip 1em plus 0.5em minus 0.4em\relax The MIT Press, 2018.

\bibitem{mnih2013}
V.~Mnih, K.~Kavukcuoglu, D.~Silver, A.~Graves, I.~Antonoglou, D.~Wierstra, and
  M.~Riedmiller, ``Playing atari with deep reinforcement learning,''
  \emph{arXiv preprint arXiv:1312.5602}, 2013.

\bibitem{Silver2016}
D.~Silver, A.~Huang, C.~Maddison, A.~Guez, L.~Sifre, G.~van~den Driessche,
  J.~Schrittwieser, I.~Antonoglou, V.~Panneershelvam, M.~Lanctot, S.~Dieleman,
  D.~Grewe, J.~Nham, N.~Kalchbrenner, I.~Sutskever, T.~Lillicrap, M.~Leach,
  K.~Kavukcuoglu, T.~Graepel, and D.~Hassabis, ``{Mastering the game of Go with
  deep neural networks and tree search},'' \emph{Nature}, vol. 529, no. 7587,
  p. 484–489, 2016.

\bibitem{Reddy2018}
G.~Reddy, J.~Wong-Ng, A.~Celani, T.~J. Sejnowski, and M.~Vergassola, ``Glider
  soaring via reinforcement learning in the field,'' \emph{Nature}, vol. 562,
  no. 7726, pp. 236--239, 2018.

\bibitem{Bellemare2020}
M.~Bellemare, S.~Candido, P.~Castro, J.~Gong, M.~Machado, S.~Moitra, S.~Ponda,
  and Z.~Wang, ``{Autonomous navigation of stratospheric balloons using
  reinforcement learning},'' \emph{Nature}, vol. 588, no. 7836, p. 77–82,
  2020.

\bibitem{Gunnarson2021}
P.~Gunnarson, I.~Mandralis, G.~Novati, P.~Koumoutsakos, and J.~O. Dabiri,
  ``Learning efficient navigation in vortical flow fields,'' \emph{Nature
  Communications}, vol.~12, no.~1, p. 7143, 2021.

\bibitem{Li2020}
B.~Li and Y.~Wu, ``Path planning for uav ground target tracking via deep
  reinforcement learning,'' \emph{IEEE Access}, vol.~8, pp. 29\,064--29\,074,
  2020.

\bibitem{Fossen2002}
T.~Fossen, ``Marine control systems: guidance, navigation and control of ships,
  rigs and underwater vehicles,'' \emph{Marine Cybernetics}, 2002.

\bibitem{Masmitja2018b}
I.~Masmitja, J.~Gonzalez, C.~Galarza, S.~Gomariz, J.~Aguzzi, and J.~Del~Rio,
  ``New vectorial propulsion system and trajectory control designs for improved
  auv mission autonomy,'' \emph{Sensors}, vol.~18, no.~4, 2018.

\bibitem{Smith2021}
\BIBentryALTinterwordspacing
K.~L. Smith, A.~D. Sherman, P.~R. McGill, R.~G. Henthorn, J.~Ferreira, T.~P.
  Connolly, and C.~L. Huffard, ``Abyssal benthic rover, an autonomous vehicle
  for long-term monitoring of deep-ocean processes,'' \emph{Science Robotics},
  vol.~6, no.~60, p. eabl4925, 2021. [Online]. Available:
  \url{https://www.science.org/doi/abs/10.1126/scirobotics.abl4925}
\BIBentrySTDinterwordspacing

\bibitem{Ryan2017}
R.~Lowe, Y.~Wu, A.~Tamar, J.~Harb, P.~Abbeel, and I.~Mordatch, ``Multi-agent
  actor-critic for mixed cooperative-competitive environments,'' in
  \emph{Proceedings of the 31st International Conference on Neural Information
  Processing Systems}, ser. NIPS'17.\hskip 1em plus 0.5em minus 0.4em\relax Red
  Hook, NY, USA: Curran Associates Inc., 2017, p. 6382–6393.

\bibitem{Terry2020}
J.~K. Terry, B.~Black, N.~Grammel, M.~Jayakumar, A.~Hari, R.~Sulivan,
  L.~Santos, R.~Perez, C.~Horsch, C.~Dieffendahl, N.~L. Williams, Y.~Lokesh,
  R.~Sullivan, and P.~Ravi, ``Pettingzoo: Gym for multi-agent reinforcement
  learning,'' \emph{arXiv preprint arXiv:2009.14471}, 2020.

\bibitem{Masmitja2016}
I.~Masmitja, O.~Pallares, S.~Gomariz, J.~D. Rio, T.~O'Reilly, and B.~Kieft,
  ``Range-only underwater target localization : error characterization,''
  \emph{"21st IMEKO TC4 International Symposium and 19th International Workshop
  on ADC Modelling and Testing Understanding the World through Electrical and
  Electronic Measurement}, pp. 267--271, 2016.

\bibitem{Memarian2021}
F.~Memarian, W.~Goo, R.~Lioutikov, S.~Niekum, and U.~Topcu, ``Self-supervised
  online reward shaping in sparse-reward environments,'' \emph{arXiv preprint
  arXiv:2103.04529}, 2021.

\bibitem{Nuzhin2021}
E.~E. Nuzhin, M.~E. Panov, and N.~V. Brilliantov, ``Why animals swirl and how
  they group,'' \emph{Scientific Reports}, vol.~11, no.~1, p. 20843, 2021.

\bibitem{Lillicrap2019}
T.~P. Lillicrap, J.~J. Hunt, A.~Pritzel, N.~Heess, T.~Erez, Y.~Tassa,
  D.~Silver, and D.~Wierstra, ``Continuous control with deep reinforcement
  learning,'' \emph{arXiv preprint arXiv:1509.02971}, 2019.

\bibitem{Fujimoto2018}
S.~Fujimoto, H.~van Hoof, and D.~Meger, ``Addressing function approximation
  error in actor-critic methods,'' \emph{arXiv preprint arXiv:1802.09477},
  2018.

\bibitem{Haarnoja2018}
T.~Haarnoja, A.~Zhou, P.~Abbeel, and S.~Levine, ``Soft actor-critic: Off-policy
  maximum entropy deep reinforcement learning with a stochastic actor,''
  \emph{arXiv preprint arXiv:1801.01290}, 2018.

\bibitem{Smith2018}
S.~L. Smith, P.-J. Kindermans, C.~Ying, and Q.~V. Le, ``Don't decay the
  learning rate, increase the batch size,'' \emph{arXiv preprint
  arXiv:1711.00489}, 2018.

\bibitem{Song1999}
T.~L. Song, ``Observability of target tracking with range-only measurements,''
  \emph{IEEE Journal of Oceanic Engineering}, vol.~24, no.~3, pp. 383--387,
  1999.

\bibitem{Jouffroy2006}
J.~Jouffroy and J.~Reger, ``An algebraic perspective to single-transponder
  underwater navigation,'' in \emph{2006 IEEE Conference on Computer Aided
  Control System Design, 2006 IEEE International Conference on Control
  Applications, 2006 IEEE International Symposium on Intelligent Control},
  2006, pp. 1789--1794.

\bibitem{Moreno2010}
D.~Moreno, A.~Pascoal, A.~Alcocer, and J.~Aranda, ``Optimal sensor placement
  for underwater target positioning with noisy range measurements,'' \emph{IFAC
  Proceedings Volumes}, vol.~43, no.~20, pp. 85--90, 2010, 8th IFAC Conference
  on Control Applications in Marine Systems.

\bibitem{Agarwal2021}
R.~Agarwal, M.~Schwarzer, P.~S. Castro, A.~Courville, and M.~G. Bellemare,
  ``Deep reinforcement learning at the edge of the statistical precipice,''
  \emph{Part of Advances in Neural Information Processing Systems 34
  pre-proceedings (NeurIPS 2021)}, 2021.

\end{thebibliography}
}

\end{document}